\documentclass{article}

\usepackage{microtype}
\usepackage{graphicx}
\usepackage{subfigure}
\usepackage{booktabs}
\usepackage{multirow}
\usepackage{pifont}
\usepackage[numbers]{natbib}

\usepackage{hyperref}
\usepackage[margin=1in]{geometry}
\usepackage{times}

\usepackage{amsmath}
\usepackage{amssymb}
\usepackage{mathtools}
\usepackage{amsthm}

\usepackage[capitalize,noabbrev]{cleveref}

\theoremstyle{plain}

\theoremstyle{definition}

\theoremstyle{remark}

\usepackage[textsize=tiny]{todonotes}

\newcommand\T{\rule{0pt}{2.6ex}}

\usepackage{array}
\newcolumntype{P}[1]{>{\centering\arraybackslash}p{#1}}
\newcolumntype{M}[1]{>{\centering\arraybackslash}m{#1}}

\begin{document}
\title{\fontsize{16.7}{16}\selectfont Learning the Language of NVMe Streams for Ransomware Detection}

\author{Barak Bringoltz, Elisha Halperin, Ran Feraru, Evgeny Blaichman, Amit Berman
\\ Samsung Semiconductor Israel Research and Development Center, Tel Aviv, Israel
\\ \{barak.b, elisha.h, ran.feraru, evgeny.bl, amit.berman\}@samsung.com
}

\date{}

\maketitle

\begin{abstract}
We apply language modeling techniques to detect ransomware activity in NVMe command sequences. We design and train two types of transformer-based models: the Command-Level Transformer (CLT) performs in-context token classification to determine whether individual commands are initiated by ransomware, and the Patch-Level Transformer (PLT) predicts the volume of data accessed by ransomware within a patch of commands. We present both model designs and the corresponding tokenization and embedding schemes and show that they improve over state-of-the-art tabular methods by up to $24\%$ in missed-detection rate, $66\%$ in data loss prevention, and $84\%$ in identifying data accessed by ransomware.

\end{abstract}

\section{Introduction}
\label{intro}
Ransomware is a cyber attack that encrypts a victim's data until a ransom is paid. It is estimated that in 2021 the total global financial loss due to Ransomware was at the 20 billion USD scale, affecting healthcare, government, law enforcement, and education \cite{Oz-et-al}, and making it a serious threat to data privacy and to national security. Ransomware detection is an active area of research in cybersecurity with detection techniques belonging to one of two types: `static' methods which identify malicious software signatures, and `behavioral' methods, which detect patterns in the behavior of computing systems infected by ransomware. Especially, machine learning techniques are becoming widely used for detection, with features generated at different levels of the software stack -- from low levels I/O to network activity \cite{Ispahany-et-al}.

\begin{figure}[t]
\begin{center}
\centerline{\includegraphics[width=\columnwidth]{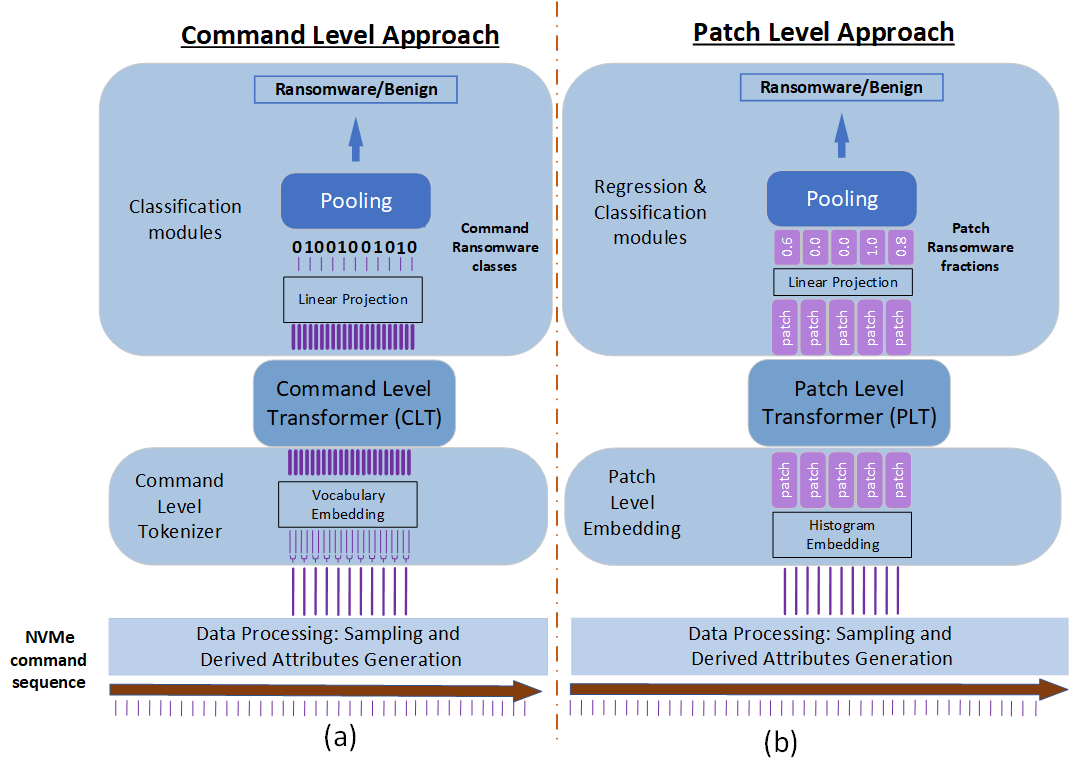}}
\caption{A high-level diagram of our ransomware detection pipelines for both the (a) CLT (left panel) and (b) PLT (right panel) models. Both first sample and process a set of commands from the NVMe stream. Then, each tokenizes the commands, feeds the tokens into its model, and outputs a prediction per token. These token predictions are pooled to produce a final binary prediction.}
\label{fig: system}
\end{center}
\vskip -0.2in
\end{figure}

Indeed, several works \cite{SSD-insider-2018, Wang-et-al-Mimosa-2019, Wang2024-alibaba} use NVMe commands to detect ransomware by aggregating them into several dozens of statistical tabular features. In this work, we take a novel approach and especially:
\begin{enumerate}
    \item\label{inov: DL} We utilize the sequential nature of the NVMe command stream, and design our features and models to incorporate the time dimension of the data.
    \item\label{inov: LM} We work with raw NVMe command sequences, without aggregating them. In particular, we introduce an appropriate tokenization scheme, enabling the model to understand commands in their context. This allows us to output predictions for individual commands/patches of commands, and recognize ransomware patterns that would otherwise be lost to aggregation.
    \item\label{inov: Performance} Finally, we show that our approach achieves superior results compared to SotA in key performance metrics such as missed detection rate and data loss prevention.
\end{enumerate}
In \cref{fig: system} we show a high-level view of our ransomware detection pipeline, and present two distinct approaches to model the problem. The first (\cref{fig: system}a) classifies individual commands into ransomware or benign, while the second (\cref{fig: system}b) works on patches of commands and predicts the fraction of ransomware they contain. As we will show%in \cref{experiments}
, both approaches perform better than current SotA, yet each has its advantages compared to the other. The first detects ransomware faster, while the second detects more ransomware overall. Thus in the following sections, we present both, along with the way they process, tokenize, and model the NVMe stream, and the viability of their implementation in a product.

\section{Data Processing}
\label{processing}
NVMe data is the set of sequential commands the Operating System (OS) sends the controller of a storage device,  instructing it what data to read or write. An NVMe command comprises of 
four fundamental attributes: the floating point  $timestamp$ denotes the time of command initialization, $opcode$ is a categorical attribute denoting the command operation (read: $R$ or write: $W$), the integer $size$ is the sum of bytes the command refers to, and the ordinal $offset$ denoting the byte offset on disk (disk addresses in the range $[offset, offset+size)$ constitute the memory space referred to by the command). 

\subsection{Derived per-command attributes}
\label{WAR}
The very nature of ransomware is to read data, encrypt it, and then destroy the original data. This led several works (e.g. \cite{Kharaz-et-al-Redemption-2017}) to calculate the overwritten byte volume and use it as a feature in detection systems. Inspired by such ideas we complement the fundamental  NVMe attributes with additional derived features. In particular, for each command $c$ with $opcode(c)=W$, we identify the command $c'$ that is the latest of all earlier commands with $opcode(c')=R$ and that has overlap $OV_{WAR}$ with the bytes of $c$ (for `write-after-read'). We also calculate the time-lapse $\Delta t_{WAR}(c)=timestamp(c)-timestamp(c')$ between $c$ and $c'$. Because we observed ransomware patterns with multiple read and write operations to the same storage location, we calculate three additional overlaps and time lapses that correspond to all other opcode combinations $(opcode(c),opcode(c'))= (R, R), (R, W), (W, W)$. Similar derived attributes were defined in \cite{Wang2024-alibaba}. Finally, we calculate the per-command time difference $\delta t(c) \equiv timestamp(c)-timestamp(c-1)$ for each command $c$ to keep track of the IO rate.

\subsection{Data Sampling}
\label{slicing}
To generate samples for train and test we sample data by slicing each NVMe stream into successive non-overlapping units of data slices. There are three natural options to define the slices, by equal (i) time extent, (ii) number of commands, or (iii) byte volume. A common choice in the literature is to use option (i) with a time extent of $\sim 10-60$ seconds (e.g. \cite{SSD-insider-2018, Hirano-et-al-2022b, Wang2024-alibaba, reategui2024_ibm}). Because the SotA models as well as some of our models can perform detection only after at least one data slice is complete, the earliest detection with option (i) at a typical I/O throughput of 130MB/sec can happen after over a gigabyte of data was read/overwritten by ransomware. To avoid this issue and the potential lack of robustness to throughput variability, we choose to sample with option (ii) and a slice size of 16500 commands or option (iii) with a slice size of 0.5GiB. For brevity, we term the models that are based on slicing options (ii) and (iii) as \texttt{ByCommand} and \texttt{ByVolume} version models.

\section{Models}
\label{models}
The structure of our problem requires us to identify the presence of malicious commands, often interspersed with differing amounts of benign commands in between. Any single command or small patch of commands in itself is very hard to model, and it is only when taken with other relevant commands that we can attempt to classify it. Attention mechanism, with its ability to identify relevant information at varying, and often long, distances, and ignore large amounts of irrelevant information in between, is a natural choice for this problem. 
\footnote{Attempts with a UNET model -- with up to x340 more parameters than the transformers -- performed worse than the PLT and CLT, so we omit it from the main paper, presenting it in \cref{appendix: models}.} The architecture itself follows the encoder parts of the architecture presented in \cite{attention}, with exact specifications given in \cref{appendix: models}. In the remainder of this section, we introduce two modeling approaches: at the command level, using the Command Level Transformer (CLT), and at the level of small command patches, using the Patch Level Transformer (PLT).

\subsection{The Command Level Transformer}
\label{CL}
The CLT -- depicted in \cref{fig: system}a -- works on frames of 250 commands, which the tokenization process turns into 500 tokens (for tokenization details see next section). The transformer is made up of three self-attention encoding layers, which follow the architecture presented in \cite{attention}. On top of the transformer, we then compose a classification module, comprised of a fully connected projection layer -- reducing the output dimension back to $500\times1$ -- and a $2\times1$ convolution with stride 2, producing 250 outputs. These are fed through a softmax layer to predict a label (ransomware or benign) per input command. Thus, the model predicts the label of each command individually, given its context in the frame. Training loss is binary cross-entropy, applied to each of the 250 predictions. During inference, the per-command predictions are averaged over a slice, and this average is then imposed by a threshold to produce a final binary prediction for the entire slice. Full details of the CLT parameters are in \cref{appendix: models}.

\subsubsection{Command Level Tokenization}
\label{CLT_tokens}
Each NVMe command contains 7 numerical attributes of different types, altogether containing up to 40 digits per command. Common tokenization methods (e.g. those presented in \cite{singh2024tokenization}) would thus result in dozens of tokens per command, and prohibitively long sequences. Instead, we first compress the commands by quantizing each attribute into a few bits, and then tokenize the compressed representation by concatenating those bits into integer tokens.
The quantization process uses domain knowledge\footnote{e.g. typical distributions of command $size$ and $\delta t$, typical file sizes and OS file system behavior for $offset$ bit choice, etc.} to preserve as much information as possible, and is described in detail in \cref{appendix: tokenization}. \cref{tab: CLT_tokenization} specifies the quantization method and the bit output for each attribute.
\begin{table}[h]
\caption{Quantization and tokenization of NVMe command. The bits are concatenated into a pair of 9-bit tokens, and an additional 10$^{th}$ bit is added to each token as its index in the pair.}
\label{tab: CLT_tokenization}
\renewcommand{\arraystretch}{1.5}
\vskip 0.15in
\begin{center}
\begin{small}
\begin{tabular}{|M{1.0cm}|M{1.1cm}|M{0.4cm}||M{1.3cm}|M{1.3cm}|M{0.4cm}|}
\toprule
     Attribute &  Operation &  Bits   &   Attribute &  Operation &  Bits\\ \hline
     $\delta t$  &  $\log{N}$ & 4 & $offset$  &   take MSB  & 4 \\ \hline
     $size$  & $\log{N}$ & 4 & $offset$ & take LSB  & 2 \\ \hline
     $opcode$  &  None & 1  & $OV_{WAR}$, $OV_{RAR}$, $OV_{RAW}$  &  binarize & 3 \\ \hline
     $index_1$  &  always 0 & 1 & $index_2$& always 1 & 1 \\
\bottomrule
\end{tabular}
\end{small}
\end{center}
\vskip -0.1in
\end{table}

A simple concatenation of the resulting 18 bits into a single token would result in a 256K vocabulary size, forcing us to hold a large (128MB) embedding table to handle it. With such a large vocabulary, many tokens are never seen during training -- over 60\% for our training data -- which may later result in unpredictable behavior during inference time, when such unseen tokens are encountered. Hence, we split each 18-bit token into two 9-bit tokens and trade effective context length, which is reduced from 500 to 250 commands, for a smaller vocabulary size. To make sure the learning algorithm treats the two tokens differently we add a 10$^{th}$ ``index'' bit to each token in the pair -- $index_1=0$ for the first and $index_2=1$ for the second -- thus fully separating the tokens in the embedding space, at the cost of increasing the vocabulary size to 1024, which still avoids the large vocabulary problems mentioned above.

\subsection{Patch Level Transformer}
\label{PL}
The PLT shown in \cref{fig: system}b follows the idea of \cite{ViT-2020} and instead of working on individual commands works on 100 small patches of the command stream -- these are the tokens of the PLT described in the next section. The transformer has six transformer encoding layers. As depicted in \cref{fig: system}b, it has a final regression and classification module comprised of a linear projection to reduce the output dimension to $100\times2$ and a sigmoid, predicting two fractions per token representing the read and write ransomware IO volumes in a patch. The model is trained to perform per-token fractional regression \cite{fractional_reg-2011} with a cross-entropy (CE) loss by summing the CE terms of all $100\times 2$ fractional volumes. To get a final prediction for an entire slice, the fractions are summed per token and the result is pooled by averaging across tokens to provide a ransomware/benign prediction probability. We threshold this probability to obtain a binary class per slice. More details on hyper-parameters are in \cref{appendix: models}.

\subsubsection{Patch Level Embedding}
\label{PLT_tokenization}
To generate tokens each slice is divided into 100 patches by sliding a window. When we model slices with a uniform number of commands (\texttt{ByCommand} slicing), we choose each patch to contain 250 commands with a stride of 165. When we model slices with uniform byte volume (\texttt{ByVolume} slicing) we choose each patch to have a total $size$ of $50MB$ with stride $\simeq 5MB$.

Distinct from the embedding method of \cite{ViT-2020} and because each command is coupled with attributes of different value types, we find that a natural way to embed the patch data is by quantizing attributes into discrete bins and calculating histograms across commands of certain types (read/write/write-after-read/read-after-read). Further details about this histogram embedding are in \cref{tab: histograms}, where we denote by $WAR$ and $RAR$ commands with $OV_{WAR}>0$ and $OV_{RAR}>0$ respectively, and by $Rest$ the commands with no such overlaps. Some of the histograms are weighted by $size$ or by $OV$. This weighting is chosen to reflect the fact that most ransomware commands carry high $size$ because they aim to read and write as fast as possible, thereby enhancing the impact of such commands.
Finally, to avoid over-fitting to patterns on certain disk locations we normalize the $offset$ values in a slice to zero mean and unit variance, and to increase robustness to throughput variability, we normalized $\delta t$ and $\Delta t$ by factors that represent their exponentially back averages across past slices. 

In addition, to easily capture information about the total byte size and amount of commands in each patch, we calculate these for read, write, $WAR$, and $RAR$ commands, as well as for all command types together. We normalize these additional 9 features to the range $[0, 1]$, and concatenate them to the rest, forming a $d_{input}=181$-dimensional embedding space per token. Further details appear in \cref{appendix: tokenization}.

\begin{table}[h]
\caption{Details of patch tokens embedding. The $offset$ and $\delta t$ histograms are weighted by $size$ and the histograms for $\Delta t_{WAR}$ and $\Delta t_{RAR}$ by $OV_{WAR}$ and $OV_{RAR}$ respectively.}
\label{tab: histograms}
\renewcommand{\arraystretch}{1.5}
\vskip 0.15in
\begin{center}
\begin{small}
\begin{tabular}{|M{1.1cm}|M{1.2cm}|M{2.7cm}|M{0.5cm}|}
\toprule
     NVMe attribute &  Histogram Weight &  Commands types & Bins\\ \hline
     $\log size$  &  - &  read/write/$Rest$ & 12 \\
     $\log OV$  &   - & $WAR$, $RAR$ & 12 \\ \hline
    \multirow{2}{1.5cm}{$offset$}  &  \quad \multirow{2}{0.8cm}{$size$} &  read/write & 14 \\
     & &  $WAR$, $RAR$, $Rest$ &  14 \\ \hline
     $\Delta t$ &   $OV$ &  $WAR$, $RAR$ & 14  \\ \hline
     $\delta t$ &   $size$ &  Any & 14 \\
\bottomrule
\end{tabular}
\end{small}
\end{center}
\vskip -0.1in
\end{table}

\section{Related work} 

\subsection{Ransomware Defense using Storage I/O Attributes}
\label{RW_w_IO}
Almost all research we are aware of combines I/O features with data at the file-system, process, and byte-data levels. For example, works like \cite{Shukla-et-al, Kharaz-et-al-Redemption-2017, Paik-et-al-2018, Amoeba-2018, Wang-et-al-Mimosa-2019} use a variety of I/O features like the number of bytes read, written, and overwritten, file-level features such as file-type and file-path diversity or patterns in directory transversal. Some also use the written byte entropy, and the similarity between the read and overwritten data. Algorithms in these earlier works were mostly rule-based algorithms (the exception is logistic regression in \cite{Amoeba-2018}). More advanced tabular methods were used in \cite{SSD-insider-2018, Baek-et-al-SSD++-2021} where IO and entropy features were calculated from time windows a few seconds wide, fed into a Decision Tree to generate predictions later aggregated along wider time windows. In \cite{Hirano-et-al-2022a} and \cite{Hirano-et-al-2022b} the authors add more I/O features and train a Random Forest (RF), an SVM, and a KNN model at granularity of tens of seconds. An RF was also used in \cite{eBPF-2023} applied to higher OS data at the process ID granularity, and \cite{Minding_the_Semantic_Gap-2024} used a Decision Tree. More recently, an XGBoost \cite{xgboost-paper}, was used in \cite{Wang2024-alibaba} to form the DeftPunk model using features like the IO size, IO byte, IOPS, and $offset$ statistics for different types of commands (read, write, overwrite, multi-read, etc.). XGBoost was also used in \cite{reategui2024_ibm} with entropy, I/O, and file-level features, calculated from a window a few seconds wide.

The data used in the majority of works involves between 1--15 families of ransomware (see \cite{Amoeba-2018, SSD-insider-2018, Wang-et-al-Mimosa-2019, Baek-et-al-SSD++-2021, Hirano-et-al-2022b, eBPF-2023, Minding_the_Semantic_Gap-2024, Wang2024-alibaba}) with the exception of \cite{Kharaz-et-al-Redemption-2017} using 29 families and \cite{TravellingHypervisor} using 32. 
The total volume read or written in \cite{Hirano-et-al-2022a} and \cite{Wang2024-alibaba} at $9$ terabyte and $12$ terabyte respectively. 
None of these published data sets is labeled at the command level making it impossible for us to use.

The SotA ransomware detection algorithms we compare to are tabular models. Due to the lack of published code, we developed a Random Forest (RF) model to represent a typical tabular approach in the literature, inspired by \cite{SSD-insider-2018}. It is designed to capture patterns in $size$, $OV_{WAR}$, and $\Delta t_{WAR}$, and uses several aggregated features per slice. The second model is DeftPunk which we also implement in code. For further details on the RF and DeftPunk, we refer to \cref{appendix: models}.

\subsection{Token Level Objective with Transformers}
\label{per_token_AI}
Our training tasks are per-token classification and regression. Similar per-token tasks can be found both in the NLP and the Computer Vision literature. Especially, Named Entity Recognition (NER) is, by construction, a token-level classification task and is closest to the task our CLT performs. To our knowledge, it was first benchmarked by a transformer in \cite{Bert-2019} via fine-tuning -- see also the recent \cite{Portuguese_NER-2020}, the review on NER in \cite{NER-review-2023}, and the recent idea in \cite{GPT-NER-2023}, where the problem is treated as a generative few shot learner. A similar task to what our PLT does is described for object recognition in \cite{AllTokensMatter-2021}. There, image tokens densely participate in the classification objective by optimizing the sum of the per-token cross-entropies and the class token cross-entropy. This approach is also natural in image segmentation as done by \cite{HSI-BERT-2019} with BERT.

\section{Experiments}
\label{experiments}
We experiment with the PLT and the CLT, comparing them to the SotA represented by two models: the RF and DeftPunk which we discuss at the bottom of \cref{RW_w_IO}.

\subsection{The Data set}
\label{data}
Our experiments are performed on a data set we collect, which we annotated as benign or ransomware at the command level. 
It contains 1,464 streams with ransomware 
generated by 137 ransomware variants from 49 ransomware families, encrypting victim data generated in-house and from the NapierOne mixed file benchmark, which is specially designed for ransomware research  \cite{Napier}. 
It also contains benign streams: in-house generated (575 streams) as well as streams downloaded from 3 subsets of the data curated by Storage Networking Industry Association (SNIA) \cite{snia-trace-block-io-4928, snia-trace-block-io-388, snia-trace-block-io-158} (686 streams). We randomly choose 2/3 of the streams for train and leave the rest for evaluation. In total, our data set contains over $5\times 10^{9}$ IO commands that reflect the processing of more than 177 terabytes of data traffic. This is significantly larger than other sets in the literature both in size and in malware coverage. For a detailed description of our data see \cref{appendix: data}.

\subsection{Performance Metrics and Evaluation Method}
\label{metrics}
We measure performance by calculating three types of metrics. The first includes the missed detection rate (MDR), the false alarm rate (FAR), and the $F1$ score: $F1=2\frac{{\rm Precision}\times{\rm Recall}}{{\rm Precision}+{\rm Recall}}$. The second involves the volume of megabytes \emph{written} by the ransomware before its first detection by the model -- used as a measure of the severity of the damage it caused before being detected. We denote this quantity by $MBD$ (MegaBytes to Detection). In particular, we measure the cumulative distribution function ($CDF$) of the $MBD$ across all evaluation streams containing ransomware and quote its third upper quantile, $MBD_{3}$. The last type of metrics consists of the percentage $P_{miss}$ of ransomware IO traffic (in megabytes) missed by the model, and the percentage $P_{err}$ of benign IO traffic (in megabytes) erroneously detected as malicious.

All the reported metrics are from the evaluation set and are calculated by cross-validation as follows: we randomly divide the NVMe streams in the evaluation set into validation/test at a ratio of 1:2, set the model thresholds on the prediction probabilities on the validation set, measure performance on the test set, and repeat this process $\times$50 times. The errors we quote represent the cross-validation $1\sigma$, and for the $MDR$ and the $FAR$, we also take into account the binomial error. All models were calibrated to the same work point by choosing the $FAR$ to correspond to an average of a single false alarm for every $50GB$ of benign IO traffic.

We compare the performance of 2 versions of models -- \texttt{ByCommand} and \texttt{ByVolume} (defined by the corresponding data slicing method discussed in \cref{slicing}). We make sure that all slices in a certain version are identically the same, allowing us to unambiguously compare the $MDR$, $FAR$, and $F1$ within each version. In addition, and since $MBD_{3}$, $P_{miss}$, and $P_{err}$, are slicing-method agnostic (as they measure volume, which is independent of slicing), we can compare their measurements across versions.

\subsection{Main Results}
\label{comparison_to_baseline}
We benchmark a total of 8 models: four from the \texttt{ByCommand} version and four from \texttt{ByVolume}. In each version, we trained four models of the following types: a PLT, a CLT, and two SotA models -- a Random Forest (RF) model and a DeftPunk model.
\cref{F1_and_MBD} reports the comparison of the $MDR$, $F1$, and $MBD_{3}$ for all models.

\begin{table}[t]
    \caption{Results for the missed detection rate ($MDR$) and the $F1$ metrics in percentages, and for the $MBD_{3}$ -- which is a primary data loss metric -- in megabytes. The two parts of the table present results for the \texttt{ByCommand} and \texttt{ByVolume} version models.}
    \label{F1_and_MBD}
    % \vskip 0.15in
    \begin{center}
        \begin{small}
            \begin{sc}
                \begin{tabular}{lccc}
                    \toprule
                    \texttt{ByCommand} & MDR & $F1$ & $MBD_{3}$ \\
                    \toprule
                    RF   & 19.7$\pm$2.1 & 89.9$\pm$1.2 & 335$\pm$\,\,\,53  \\
                    DeftPunk & 16.5$\pm$1.8 & 90.8$\pm$1.0 & 366$\pm$\,\,\,58 \\ 
                    PLT  & 14.9$\pm$1.7 & 91.8$\pm$0.9 & 207$\pm$100 \\
                    CLT  & {\bf 12.5$\pm$1.1} & {\bf 93.2$\pm$0.7} & {\bf 107$\pm$\,\,\,\,\,\,5}  \\
                    \hline \T
                    \texttt{ByVolume} & MDR & $F1$ & $MBD_{3}$ \\
                    \midrule
                    RF  & 22.6$\pm$2.3 & 87.1$\pm$1.4 & 286$\pm$33  \\
                    DeftPunk & 20.4$\pm$2.0 & 88.4$\pm$1.2 & 282$\pm$34 \\ 
                    PLT & 17.4$\pm$1.2 & 90.3$\pm$0.7 & 157$\pm$18 \\
                    CLT & {\bf 16.5$\pm$1.2} & {\bf 90.8$\pm$0.6} & {\bf \,\,\,97$\pm$12}  \\
                    \hline
                \end{tabular}
        \end{sc}
        \end{small}
    \end{center}
    \vskip -0.1in
\end{table}

\begin{table}[t]
\caption{Measurements of $P_{miss}$ and of $MBD_{3}$ for the optimal models we benchmarked.}
\label{p_miss_and_MBD}
\vskip 0.15in
\begin{center}
\begin{small}
\begin{sc}
\begin{tabular}{llcc}
\toprule
Model & Version & $P_{miss}$ & $MBD_{3}$ \\
\midrule
RF    & \texttt{ByVolume} & 9.0$\pm$1.8 & 	286$\pm$33  \\
DeftPunk  & \texttt{ByVolume} & 6.7$\pm$1.1 & 282$\pm$34  \\
PLT & \texttt{ByVolume}  & {\bf 1.1$\pm$0.2} & 157$\pm$18 \\
CLT  & \texttt{ByCommand} & 2.0$\pm$0.7& {\bf \,\,\,\,97$\pm$12}	 \\  
% \hline
\bottomrule
\end{tabular}
\end{sc}
\end{small}
\end{center}
\vskip -0.1in
\end{table}

We choose the optimal model from each type according to the values of the $MBD_{3}$ and obtain the \texttt{ByCommand} CLT and the \texttt{ByVolume} RF, DeftPunk and PLT. We find that the $P_{err}$ for all these is consistent at $P_{err}=0.6\%-0.9\%$ with an error of $0.2\%-0.3\%$ and present $P_{miss}$ and $MBD_3$ in \cref{p_miss_and_MBD}. 

From \cref{F1_and_MBD} we see that the two tabular models have similar performance with DeftPunk slightly better than the RF in most metrics and moderately better in $P_{miss}$. We therefore compare the CLT and PLT to DeftPunk henceforth. We find that the CLT is the superior model in all metrics for both model versions: it is better than DeftPunk by up to $24.4\%\pm11.0\%$ in $MDR$, up to $5.8\pm1.3\%$ in $F1$, and up to $73.5\pm5.3\%$ in $MBD_{3}$. The PLT is also superior to DeftPunk by up to $7.0\pm1.2\%$ in $F1$, and up to $44.3\pm9.3\%$ in $MBD_{3}$. The PLT also has a better (lower) $MDR$ score compared to DeftPunk but it is below the statistical error. In terms of $P_{miss}$ we find that the PLT is superior to all models and that it is $84.3 \pm 4.1 \%$ better than DeftPunk while the CLT is $70.0 \pm 11.0 \%$ better.

\begin{figure}[t]
    \begin{minipage}{0.45\textwidth}
        \vskip -0.2in
        \centering
        \includegraphics[width=\textwidth]{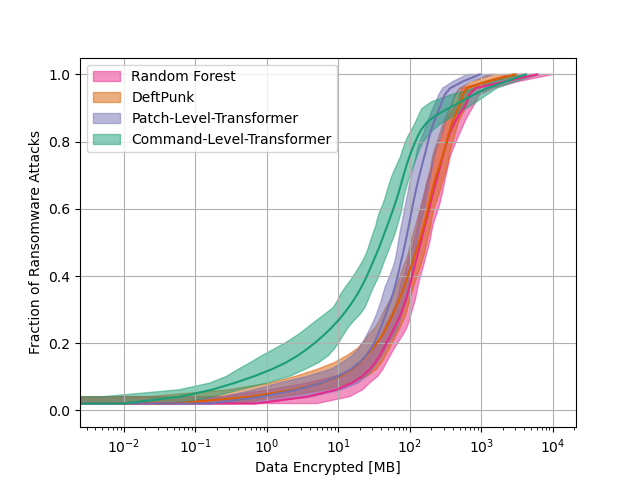}
        \caption{The cumulative distribution function of the $MBD$ for the four best models that appear in \cref{p_miss_and_MBD}, with their $1\sigma$ spread.}
        \label{fig: CDF_MBD}
        % \vskip -0.2in
    \end{minipage}
    \hfill
    \begin{minipage}{0.45\textwidth}
        % \vskip 0.2in
        \centering
        \includegraphics[width=\textwidth]{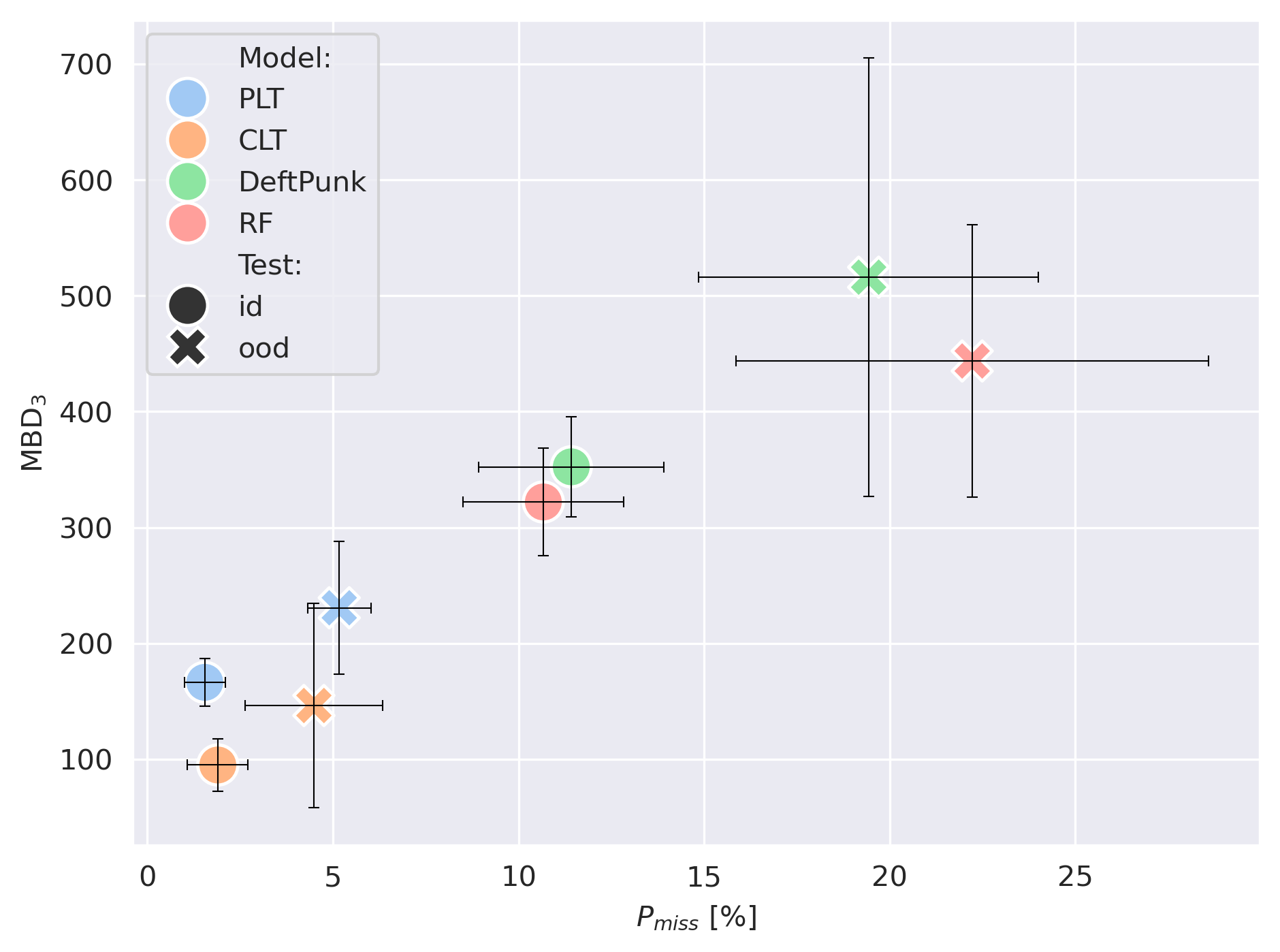}
        \caption{The in-distribution (id) and out-of-distribution (ood) results for $MBD_3$ and $P_{miss}$. The error bars take into account both results variability between folds and individual fold uncertainty.}
        \label{fig: robustness}
        % \vskip -0.2in
    \end{minipage}
\end{figure}

Finally, we plot the $CDF$ of the $MBD$ in \cref{fig: CDF_MBD} for the models in \cref{p_miss_and_MBD}.
The improvement in $MBD_{3}$ is also clear if we inspect \cref{fig: CDF_MBD} where we see that the CLT is very good in terms of detecting ransomware fast, but is prone to a non-negligible tail at very high $MBD$ values. In contrast, the PLT can prevent these situations but is not as good as the CLT at low values of $MBD$. Both models are superior to the tabular models which are themselves similar.

\subsection{Robustness to Unseen Ransomware}
\label{robustness}
We now analyze the ability of our models to generalize to unseen ransomware. Since differently named ransomware may in actuality be similar in their behavior, we would like to ensure that we do not have data leaks due to similar ransomware misclassified as unseen. To achieve this we first perform a clustering analysis of the 137 ransomware variants we have in our data set. As a distance metric we use TLSH \cite{TLSH} with a similarity score of 100 \cite{liu2023evaluation}. The analysis shows that the actual number of independent ransomware types in the data set is 47, of which 29 are singleton clusters and 9 contain 5 or more ransomware variants. The full list of clusters appears in \cref{appendix: data}.

We now divide the ransomware variants into 3 roughly equal groups and perform 3 test iterations. For each iteration, we mix two groups and evenly split them into train and in-distribution test sets. The third group is the out-of-distribution test set. The benign streams train and test data remain the same throughout, to provide the same negative samples to the different id and ood tests. We train the models on the train set and test on the id and ood test sets. Finally, we average the results of all three runs and present them in \cref{fig: robustness}.

We see that while there is a visible and statistically significant performance degradation on the ood set, for all models, the PLT and CLT models still perform better than the SotA. In fact, we observe that the ood performance of our models is still better than the id performance of the SotA.

\subsection{Per-Token Accuracy}
\label{accuracy}
\begin{figure}[!b]
    \begin{minipage}{0.45\textwidth}
        \vskip 0.2in
        \centering
        \includegraphics[width=\textwidth]{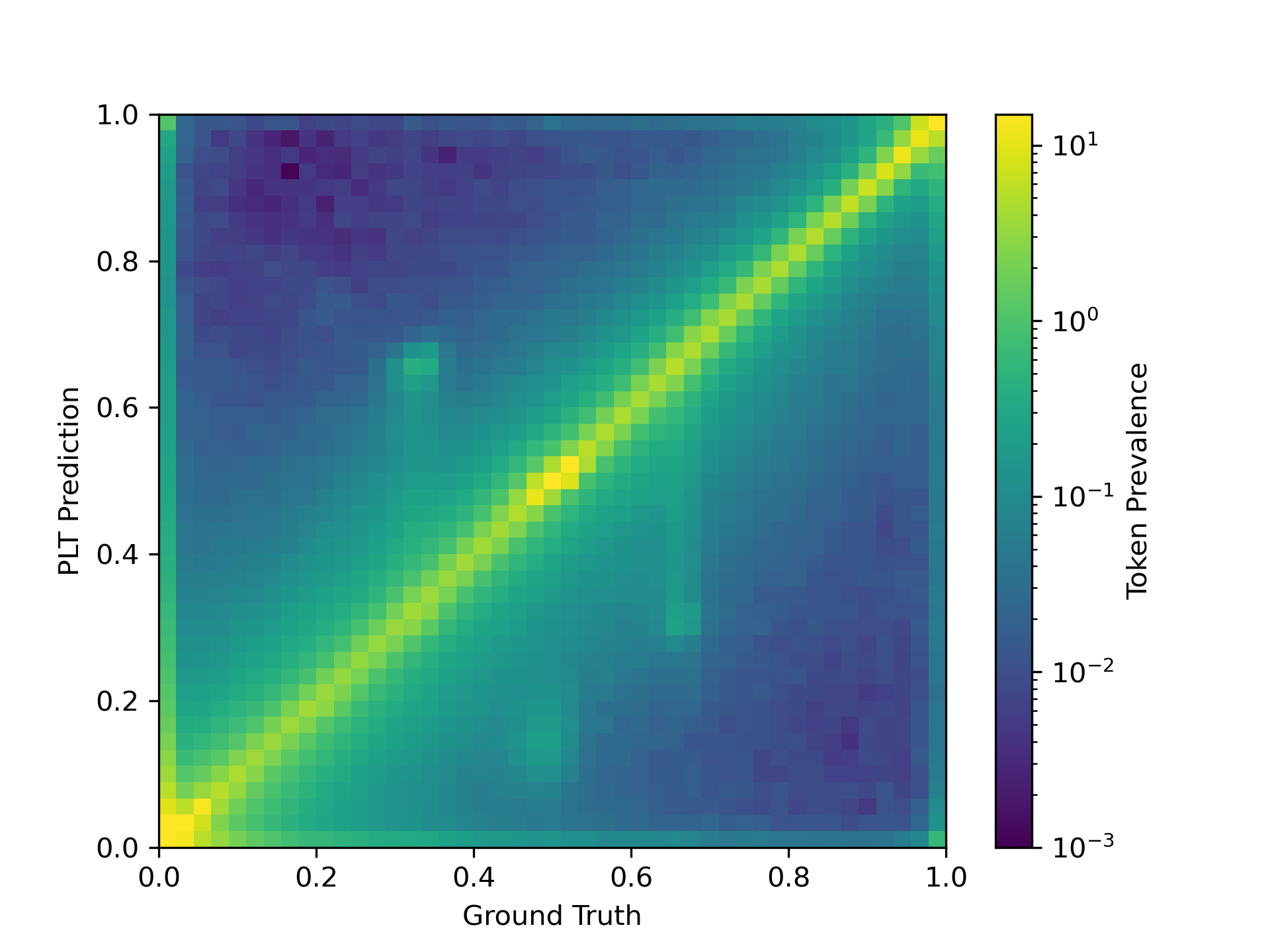}
        \caption{The histogram heatmap for the correlation between the actual fraction of ransomware IO volume and its PLT prediction.}
        \label{fig: WRITE_accuracy_PLT}
    \end{minipage}
    \hfill
    \begin{minipage}{0.45\textwidth}
        \centering
        \includegraphics[width=\textwidth]{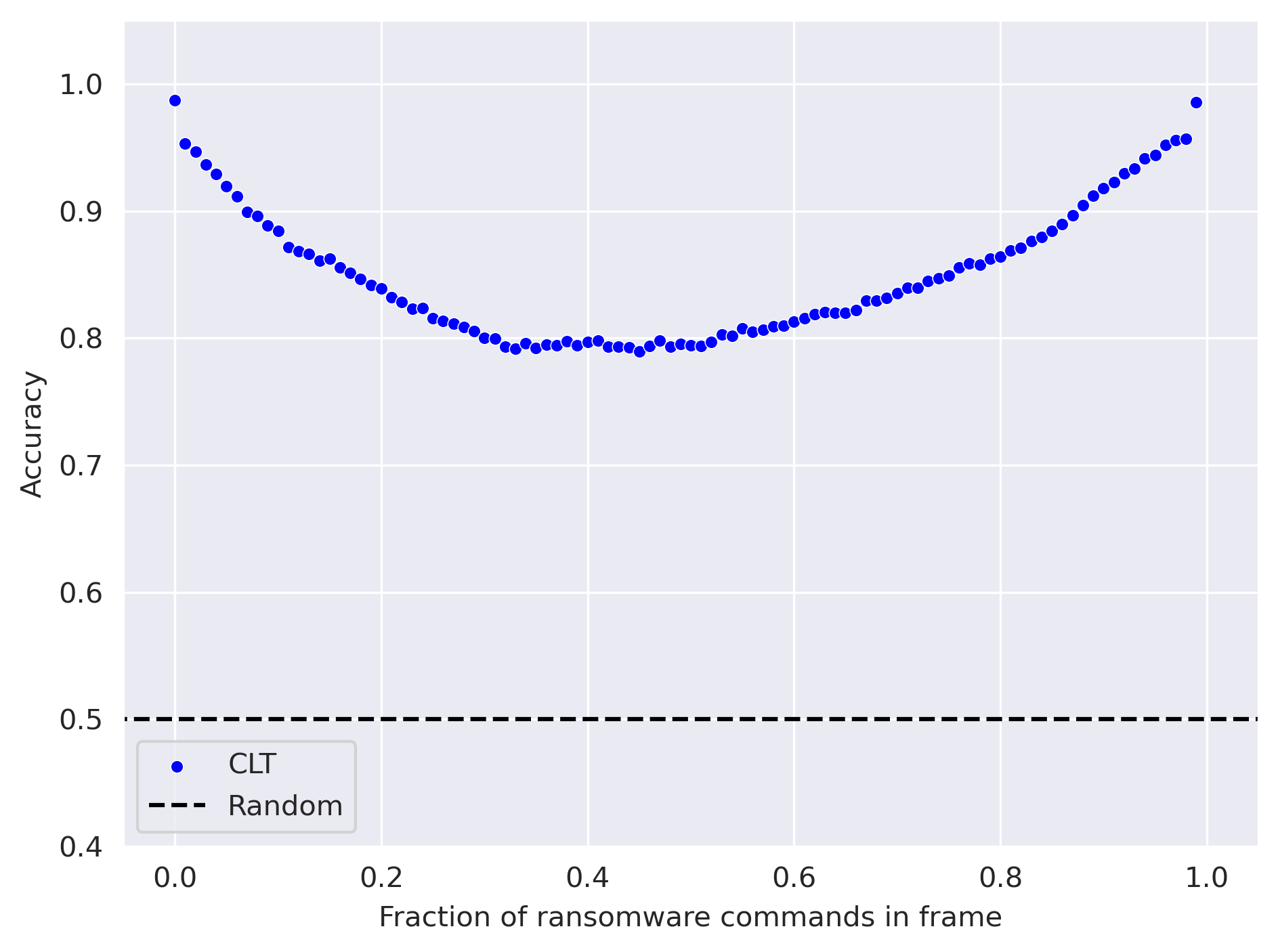}
        \caption{The accuracy of data slices for the CLT versus the slice ransomware command fraction.}
        \label{fig: accuracy_CLT}
    \end{minipage}
\end{figure}

We also find that the results within each data slice are accurate. For example, the PLT-predicted per-token read and write volume of data that is accessed by ransomware exhibits excellent accuracy -- see \cref{fig: WRITE_accuracy_PLT} for the cross-token histogram of the correlation between the per-token IO volume written by ransomware and its PLT-predicted value. When we calculate the accuracy of the PLT prediction we find the error in read and write volumes of data accessed by ransomware is at most 4.3MB and 5.5MB 
for $95\%$ of the $110K$ ransomware PLT tokens in the evaluation set.

Good within-slice accuracy is also seen for the CLT. To demonstrate this we calculated the prediction accuracy per slice (the fraction of commands in a slice whose label -- ransomware or benign -- is predicted correctly) and plot it in \cref{fig: accuracy_CLT}. We group the slices by the actual fraction of ransomware commands they contain and measure the model's average accuracy on each such group. From the plot, we see that the model accurately predicts, with $\ge 80\%$ accuracy, even for the most challenging slices whose ransomware fraction is $\sim 0.5$, i.e. that contain about equal amounts of benign and ransomware commands, interspersed among each other. Nevertheless, the model can still separate the mix into ransomware and benign commands with significantly better than random accuracy, showing command-level pattern recognition capabilities.
These analyses support our claim that the models do indeed learn low-level patterns in the data, and do not simply revert to some global aggregated function applied equally to all of their outputs.

\begin{figure}[t]
    \begin{minipage}{0.45\textwidth}
        \centering
        \includegraphics[width=\textwidth]{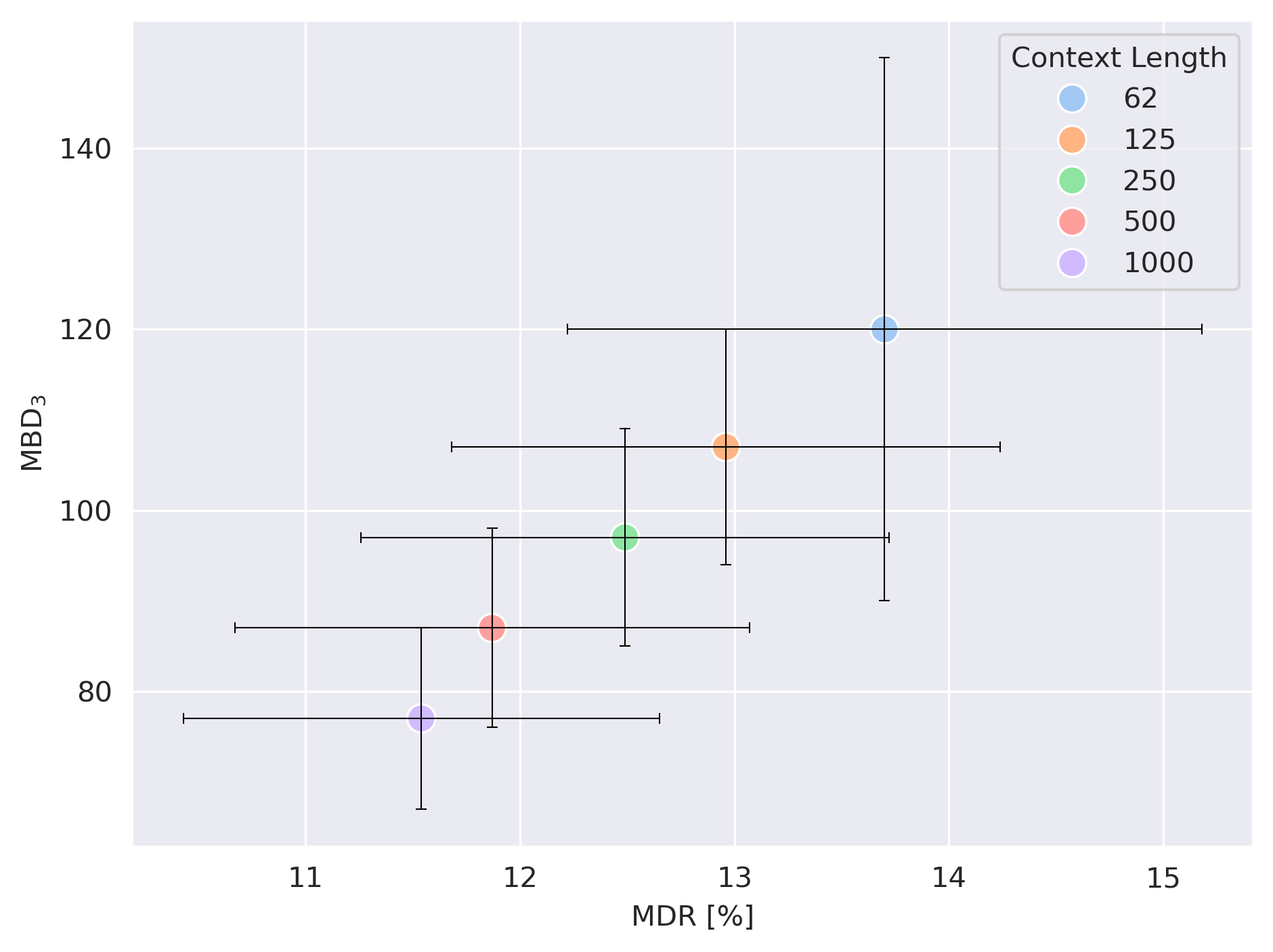}
        \caption{The dependence of $MBD_3$ and $MDR$ in of the CLT model on the context length. Note that due to splitting each command into two tokens, the actual model input length is twice the numbers shown above.}
        \label{fig: Context}
    \end{minipage}
    \hfill
    \begin{minipage}{0.45\textwidth}
        \centering
        \includegraphics[width=\textwidth]{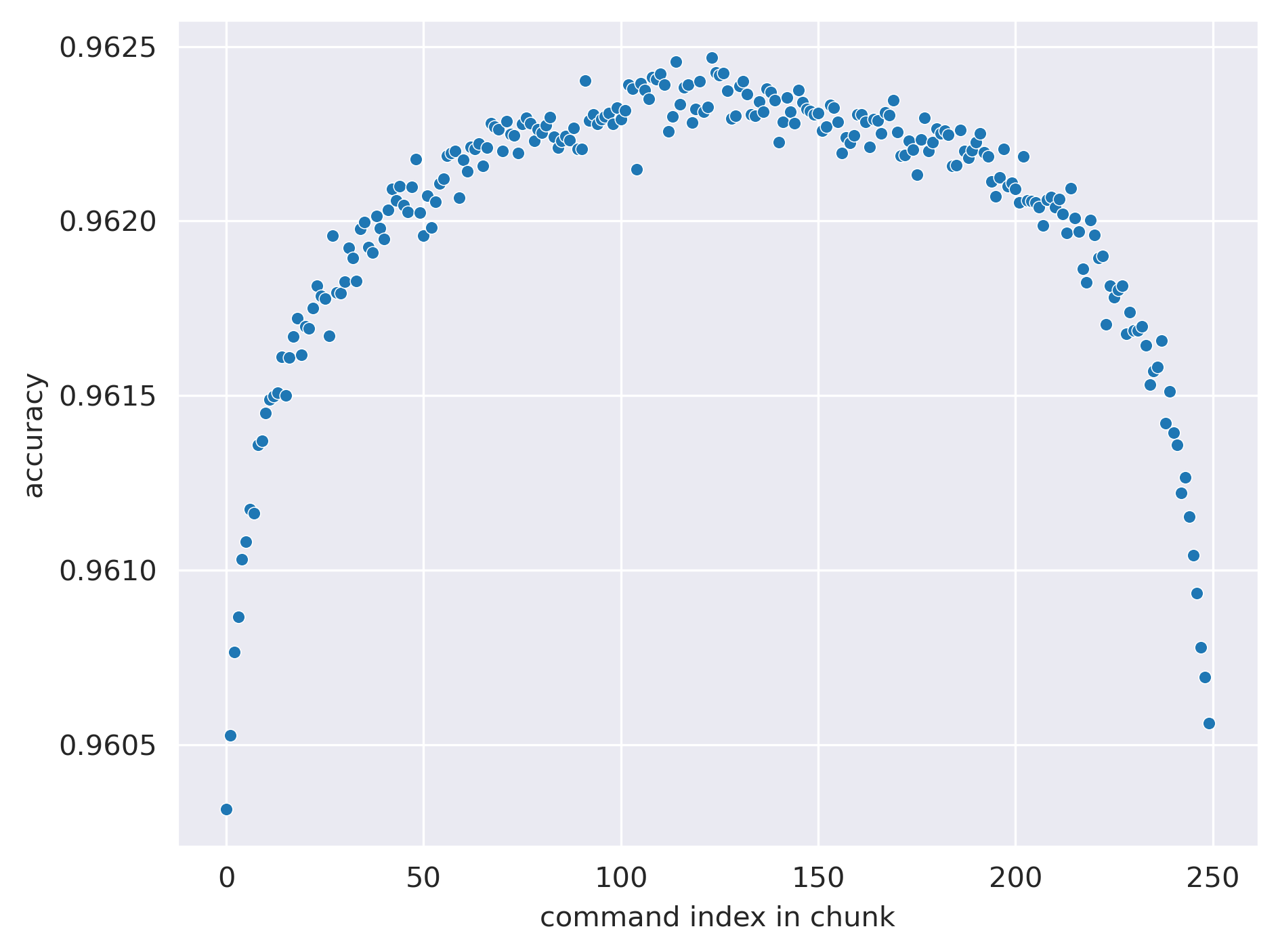}
        \caption{The dependence of command prediction accuracy on its location in the chunk. Commands in the center of the chunk have better prediction accuracy than those on either side of it.}
        \label{fig: Context_chunk}
    \end{minipage}
\end{figure}

\subsection{The Importance of Context}
\label{context}
As discussed above, it is very unlikely that a single command can be modeled well on its own since only its appearance within the context of other commands can determine its origin. Thus, providing sufficient context was an important consideration in our transformer-based architectural choice for the CLT, and here we explicitly check this line of thought. In particular, we trained a set of models with increasingly large contexts and measured their performance. We plot the results in \cref{fig: Context} where it is clear that as the context increases, the performance of the CLT improves. Since the prediction of the CLT is per-command, this means that as the context each command ``sees'' increases, so does the model's ability to correctly predict it. We also note that the CLT model variant we chose -- the one trained with $context=250$, and whose results are presented in \cref{comparison_to_baseline} -- is not the best in terms of $MBD_3$ and $MDR$. Nevertheless, as mentioned below in \cref{hw-implementation} this model was chosen as it is small enough to make it possible to implement in a product.

In \cref{fig: Context_chunk} we plot the command prediction accuracy as a function of its position in the chunk (calculated over all chunks in the test set). We see that commands in the center of the chunk -- i.e. those having at least some context on both sides -- are somewhat better predicted on average than those on either side, who have only one-sided context to draw from. While the difference is not large, working on overlapping chunks and taking only the middle $75\%$ of predictions may still improve the overall model performance.

\vspace{-0.3cm}

\subsection{Feature Ablation}
\label{ablation}
To measure the contribution of our features and tokenization to the overall performance of our models, we measure it for ablated versions of our models' token space. We do so by training a new set of PLT and CLT models while dropping sub-spaces of the embedding space as per their description in \cref{CLT_tokens} and \cref{PLT_tokenization}. The results of this study are presented in \cref{tab: token_ablation}. We also test the effect of splitting each of the NVMe commands into two CLT tokens, by training the CLT model on single tokens per command and measuring the results. We find a slight $\times1.06$ degradation in $MBD_3$ and the same performance in $P_{miss}$, assuring us that splitting the command to reduce the vocabulary size does not incur a performance degradation.

\begin{table}[h]
\caption{The results of feature ablation on the CLT and PLT models. Each feature subset was ablated together. The numbers represent the amount by which the corresponding metric increased.}
\label{tab: token_ablation}
\renewcommand{\arraystretch}{1.5}
\vskip 0.15in
\begin{center}
\begin{small}
\begin{tabular}{|M{1.8cm}|M{0.6cm}|M{0.7cm}|M{1.1cm}|M{0.6cm}|M{0.7cm}|}
\toprule
     \multicolumn{3}{|c|}{Command Token Ablation} &  \multicolumn{3}{|c|}{Patch Token Ablation} \\ \hline 
     \multirow{2}{1.5cm}{\,\,Feature Subset} &  \multicolumn{2}{c|}{Degradation} &  \multirow{2}{1.5cm}{\,\,Feature Subset} &  \multicolumn{2}{c|}{Degradation} \\ \cline{2-3} \cline{5-6}
       & $P_{miss}$ & MBD$_3$ &   &  $P_{miss}$ & MBD$_3$ \\ \hline 
     \bottomrule
     $offset$ & x2.1 & x1.6 & $size$ & x4.6 & x2.3 \\
     $\delta t$ & x1.7 & x1.3 & $\delta t$ & x5.1 & x2.2 \\
     $opcode$ & x1.4 & x1.2 & fractions & x1.4& x1.2 \\
     $size$ & x1.3 & x1.1 & $\Delta t$ & x1.2 & x1.2 \\
     $OV$ & x1.3 & x1.1 & $offset$ & $<$1.1 & $<$1.1 \\
     $index$ & x1.1 & x1.1 & $OV$ & $<$1.1 & None \\
     \bottomrule
\end{tabular}
\end{small}
\end{center}
\vskip -0.1in
\end{table}

\section{Hardware Implementation}
\label{hw-implementation}
Our model's end goal is hardware implementation, especially in SSD devices.
Thus, careful considerations must be made to ensure our algorithms can indeed meet the area-on-silicon and throughput requirements of an actual system. As shown in \cref{fig: Context}, we deliberately chose a smaller, weaker model for the CLT, to make hardware implementation feasible. We discuss the hardware requirements of the full model in \cref{appendix: HW} and below we summarize our findings.

The first requirement one needs to comply with is low DRAM usage, as determined by the number of parameters each model has. The total number of parameters in the CLT is $\sim450K$ which means it takes up $\sim1MB$ of DRAM. The PLT is a larger model, with 20M parameters, taking up 40MB of DRAM. Both these DRAM sizes can be accommodated on the internal DRAM of the SSD controller.

The second requirement is that the detection model will be fast enough to keep up with the throughput of the SSD, and we now show that this is indeed the case for our models. Starting with the CLT, we first estimate the number of multiplications it performs in each forward pass at $300M$ (see \cref{appendix: HW}). Next, assuming that we use $256$ multipliers and a clock cycle of $300$MHz we observe that a single CLT forward pass takes $\sim 300M/256/300{\rm MHz}\simeq 4$ milliseconds. Because such a CLT forward pass processes 250 commands we find that at a typical sequential workload of 256KB per command the corresponding data traffic amounts to 
$250\cdot256K\sim64$MB. Hence we see that the CLT can support a throughput of $\sim 64MB/4{\rm millisecond}=16$GBps, well above the throughput of even the fastest SSD drives currently on the market \cite{ssd-comparison-site}. 
The corresponding calculation for the PLT proceeds similarly. First, being a larger model it uses 1.8B multiplications in a single forward pass (again see \cref{appendix: HW}), taking up $24$milliseconds. Next, the fact that the PTL operates on data slices of $0.5$GiB each, means it supports throughput of $0.5\text{GiB}/24{\rm milliseconds}=21$GBps, again posing no constraint on the implementation.

Finally, since our networks work with half-precision floating point operations, and the area cost of one half-precision multiplier is $\sim2K$ gates, our total area cost is $\sim256\cdot2K=512K$ gates, which is an acceptable addition to a modern SSD controller.

\subsection{Hardware Versus Software Implementation}
\label{hw-vs-sw}
In choosing to implement the ransomware detection module into the hardware of the SSD controller we are motivated by several advantages, compared to software implementations. 

The first advantage is high computational efficiency. As shown in the previous section, our use of hardware parallelization allows us to keep up with the required SSD throughput. In contrast, we observe that processing in software with a dedicated CPU core operating at 3GHz would lead to an approximate $\times25$ slow-down compared to the hardware implementation, and to significant latency effects.
A second advantage is resistance to malware interference: while software-based solutions may be disabled by the attacking malware, solutions implemented into the SSD controller cannot be affected by such external software processes, and so are immune to attacks, and cannot be turned off or disrupted by malware.
A third advantage of our SSD-based solution is that its garbage collection naturally allows one to recover lost data even when it has been deleted at the software level, thus enabling full data recovery even for data corrupted by the ransomware before it was detected.

These advantages and the fact that any other hardware-based solutions require the addition of specialized hardware modules to the storage system -- increasing form factor, cost, and complexity -- make our implementation into the hardware of the SSD controller a very natural and comprehensive solution.

\section{Limitations}
\label{limitations}
The large amount of calculations our models perform increases their on-chip area requirements compared to SotA models such as the RF or DeftPunk. This entails that for very low-resource systems (e.g. phones), our models would require additional work to be competitive with current SotA models. Thus our product is currently intended for medium to high-resource systems (e.g. SSDs and data storage servers), and future work will improve our models to support low-resource systems as well.

\section{Summary}
\label{summary}
In this paper, we take a novel approach to the problem of detecting ransomware activity by inspecting the NVMe IO stream. We reformulate the problem, moving away from aggregative statistical features and instead work with the NVMe stream directly, making use of its sequential nature. For this purpose, we first devise a method of converting our raw data into tokens and then employ the transformer architecture which, as we show, is uniquely suited to the specific characteristics of our data. We further demonstrate that our models learn to classify individual commands in their context and measure ransomware activity within small patches of commands, even when presented with a difficult mixed stream. We also show that context matters -- the more context a command ``sees'', the better its prediction will be.

We compare our approach against SotA methods using classification metrics as well as metrics designed to measure ransomware damage, and see that our models' performance is superior to theirs and that this advantage is robust; preserved even when our models are tested on ransomware variants that are out of distribution and never seen during training. We also show that we could further improve the performance of our model by increasing its context length, but by taking instead the variant we presented, our solution becomes viable for inclusion in SSD products.

% \section*{Acknowledgements}

\bibliography{refs}
\bibliographystyle{hunsrtnat}

\newpage
\appendix
\onecolumn
\section{Model Specifications}
\label{appendix: models}
\subsection{Random Forest Model}
The first model we use as a baseline comparison for our approach is a Random Forest (RF) with 23 tabular features extending past works on decision trees and random forest approaches like \cite{SSD-insider-2018}. Especially, to avoid over-fitting and being constrained by the hardware implementation, our model is restricted to 20 trees with a maximum depth of 20. After training (using the Gini criterion) we validated through a feature importance analysis the common wisdom, which suggests the read and overwrite fractional volumes are the most important features (during a ransomware cyber attack, the read, write, and overwrite volumes are nearly identical, making the fractional read and overwrite volume very close to $1/2$). We find that the space on the disk the trained model takes up is roughly $18MB$ corresponding to 260K nodes in the forest.

\subsubsection{Features}
We recall that the featurization process is preceded by slicing the data into slices of identical read and write volume (option (iii) described in \cref{slicing}) and that we choose the slice volume to be equal to $V_0=0.5GiB$. We extract the features we present in \cref{rf-features} from each slice. In that table, we denote by $V$ the total number of logical blocks in the slice. The value of $V$ obeys $V\lesssim  V_0$ because there is no guarantee the cumulative sum of $size$ along the trace will be evenly commensurate with $V_0$. Finally by $\sigma_{OV_{WAR}}$ and $\mu_{OV_{WAR}}$ we denote the standard deviation and the mean of $OV_{WAR}$ across the commands of the slice. We choose $10$ bins for the histograms $\mathrm{H}_{R}$ and $\mathrm{H}_{WAR}$ described in \cref{rf-features} and so have $23$ features per slice.
\begin{table}[h]
\caption{RF Features}
\label{rf-features}
\renewcommand{\arraystretch}{2}
\vskip 0.15in
\begin{center}
\begin{small}
\begin{tabular}{|c|c|} \hline
$f_R$ & $\displaystyle \frac1V\sum_{c \atop opcode=R} size(c)$  \\ \hline
$f_{WAR}$ & $\displaystyle \frac1V\sum_c OV_{WAR}(c)$  \\ \hline
$\mathrm{CV}_{WAR}$ & $\sigma_{OV_{WAR}}/\mu_{OV_{WAR}}$ \\ \hline
$\mathrm{H}_{R}$  & The histogram of read commands' $size$\\ \hline
$\mathrm{H}_{WAR}$  & The histograms of $OV_{WAR}$ values\\ \hline
\end{tabular}
\end{small}
\end{center}
\end{table}

\subsection{DeftPunk Model}
The second baseline model we use is the DeftPunk model \cite{Wang2024-alibaba}. It is a two-layer model: the authors choose to apply a Decision Tree (DT) on a first set of features and pass the ransomware-suspicious samples to a second layer chosen to be XGBoost which is applied on a larger set of features. The partition between features in the first and second layers is done by the computational effort made in feature extraction and we follow the same choice described by the authors. The models we trained have a maximum depth of 6 in both the DT and the XGBoost. We found that the model on disk takes up $0.5MB$, corresponding to its 10K nodes, and 100 trees in the XGBoost.
\subsection{CLT and PLT Models}
We describe the hyperparameters chosen for the CLT and the PLT in \cref{clt-hyperparams} and \cref{plt-hyperparams}, respectively.
\begin{table}[h]
\caption{CLT Hyperparameters.}
\label{clt-hyperparams}
\renewcommand{\arraystretch}{1.5}
\vskip 0.15in
\begin{center}
\begin{small}
\begin{tabular}{|c|c|} \hline
Hyperparameter Name & Chosen Value \\ \hline
Vocabulary Size & 1,024 \\ \hline
Batch Size & 64 \\ \hline
Embedding Dimensions & 128 \\ \hline
Feedforward Dimensions & 128 \\ \hline
No. of Heads & 4 \\ \hline
No. of Layers & 3 \\ \hline
Context Length & 500 \\ \hline
Convolution Kernel Size & 2 \\ \hline
Dropout & 0.1 \\ \hline

\end{tabular}
\end{small}
\end{center}
\end{table}

\begin{table}[h]
\caption{PLT Hyperparameters}
\label{plt-hyperparams}
\renewcommand{\arraystretch}{1.5}
\vskip 0.15in
\begin{center}
\begin{small}
\begin{tabular}{|c|c|} \hline
Hyperparameter Name & Chosen Value \\ \hline
Input Size & 181 \\ \hline
Batch Size & 256 \\ \hline
Embedding Dimensions & 512 \\ \hline
Feedforward Dimensions & 2,048 \\ \hline
No. of Heads & 4 \\ \hline
No. of Layers & 6 \\ \hline
Context Length & 100 \\ \hline
Dropout & 0.1 \\ \hline

\end{tabular}
\end{small}
\end{center}
\end{table}
For both models, we used an ADAM optimizer with a learning rate of $1e^{-4}$. The CLT's optimizer used an LRScheduler of 30 steps and $\gamma=0.8$. The CLT and PLT were trained for 300 and 400 epochs, respectively.
Our models' receiver operating characteristic (ROC) thresholds, as calculated on the cross-validation set, are presented in \cref{roc-thresh}.
\begin{table}[h]
\caption{Models' ROC Thresholds}
\label{roc-thresh}
\renewcommand{\arraystretch}{1.5}
\vskip 0.15in
\begin{center}
\begin{small}
\begin{tabular}{|c|c|} \hline
Model Name & ROC Threshold \\ \hline
CLT & $0.17253 \pm 0.03187$ \\ \hline
PLT & $0.136298 \pm 0.024286$ \\ \hline
RF & $0.903377 \pm 0.017638$ \\ \hline
DeftPunk & $0.964811 \pm 0.009038$ \\ \hline

\end{tabular}
\end{small}
\end{center}
\end{table}

\subsection{UNET}
\label{appendix: UNET}
To test the importance and fit of the transformer architecture we tested other architectures like fully connected applied to a concatenation of the PLT features of all tokens, as well as 1D convolutional networks. 
The best models of these architectures were of the UNET type
with the architecture \cref{Unet_architectures}, input with the tokens of the PLT, hence $T(0)=T(f)=100$ and $F(0)=181$ and $F(f)=2$. The results of the best UNET model are presented in \cref{UNET-results} together with the result of our PLT and CLT from \cref{experiments}. The results demonstrate that the UNET is of a degraded performance compared to the transformer, even for very large models with over x10 parameters.
\subsubsection{UNET Architectures}
\label{Unet_architectures}
\begin{figure}[ht]
\vskip 0.2in
\begin{center}
\centerline{\includegraphics[width=\columnwidth]{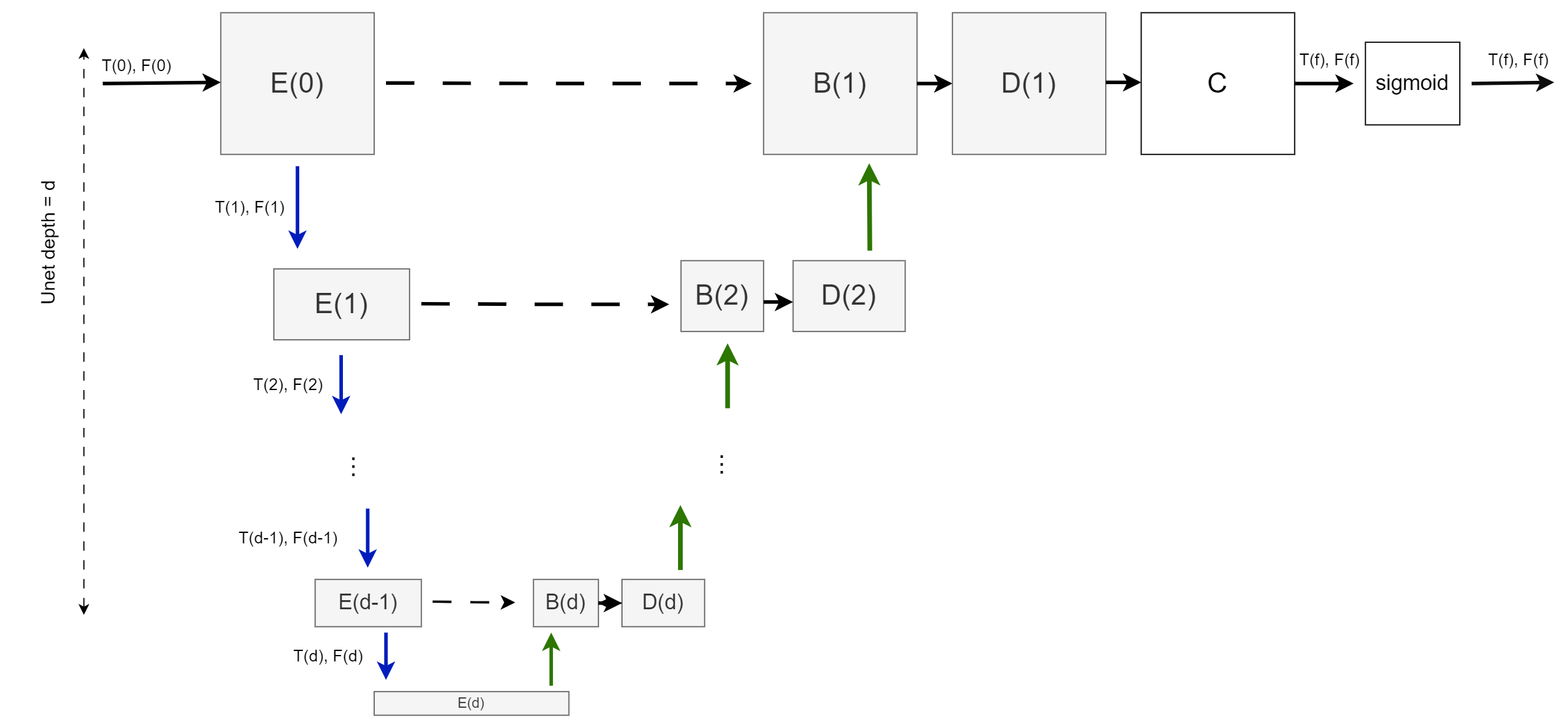}}
\caption{A pictorial representation of the UNET architecture that we use.}
\label{fig: unet_architecture}
\end{center}
\vskip -0.2in
\end{figure}

We pictorially present the specific implementation of the architecture we use in \cref{fig: unet_architecture}. 
By $T(i)$, $F(i)$ we denote the number of tokens and the embedding size respectively of the UNET level $i$. The blue arrows denote a max pooling with kernel size $= 2$ which means that $T(i+1)=T(i)//2$. The green arrow denotes transpose convolution with stride $= 2$ and with kernel size equal to $max(3, k)$, where $k$ is a parameter of the encoder and decoder blocks $E$ and $D$ that we discuss below. This transpose convolution also reduces the embedding dimension $\times 2$. The blocks $B(i)$ denote the concatenation of the UNET skip connection (the dashed arrow) and the result of the transposed convolution along the feature dimension. Finally, the block $C$ denotes a $1\times1$ convolution from an initial dimension (which after $D(1)$ is equal to $2F(0)$) into a final dimension equal to $2$, followed by a sigmoid. This provides an output of dimension $T(0)\times 2$ of elements, each bounded in the range $[0, 1]$, that is then compared to the read-and-write ransomware fractions series using a binary cross entropy loss function.

The $padding$ and $output\_padding$ parameters of the transpose convolutions depend on the value of $T(i)$ (or more precisely on whether $T(i)$, which is determined by applying pooling on the higher-level series with size $T(i-1)$, obeys $T(i-1)=T(i)\times 2$ or $T(i-1)=T(i)\times 2+1$). The main UNET parameters are the depth $d$ of the UNET and the kernel size $k$ of the $E$ and $D$ blocks in \cref{tab: unet_blocks} which we grid search for $d=0, 1, 2, 3$ and $k=3, 5, 7, 9$ (when the UNET depth is zero, $d=0$, the architecture represented by the set of gray rectangles in \cref{fig: unet_architecture} reduces to $E(0)$). The initial dimensions $T(0)$ and $F(0)$ can also be considered as hyper-parameters, but $T(f), F(f)$ are not as we fix them to obey $T(f)=T(0)$ and $F(f)=2$. We attempted various types of encoder and decoder blocks  (the $E$ and $D$ blocks in \cref{fig: unet_architecture}) including ResNet blocks and simple convolutions, and found that the best are those presented in \cref{tab: unet_blocks}.

\newpage

\begin{table}[t]
\caption{The encoder block $E$ and decoder block $D$.}
\label{tab: unet_blocks}
\renewcommand{\arraystretch}{1.5}
\vskip 0.15in
\begin{center}
\begin{small}
\begin{sc}
\begin{tabular}{|c|c|}
\toprule
 Encoder & Decoder \\ \hline
\includegraphics[width=0.4\textwidth]{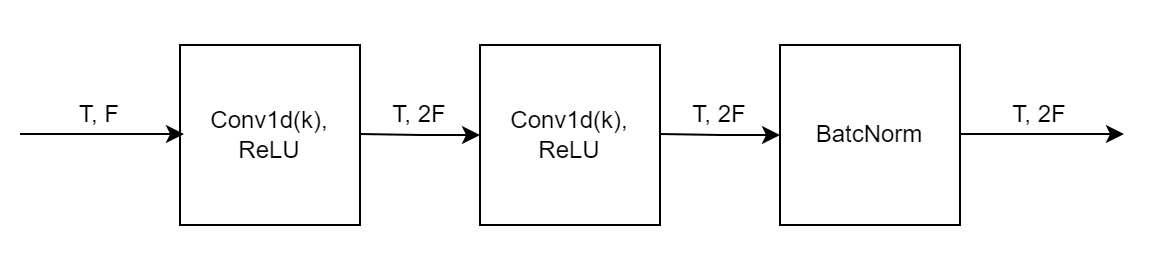}  
& 
\includegraphics[width=0.4\textwidth]{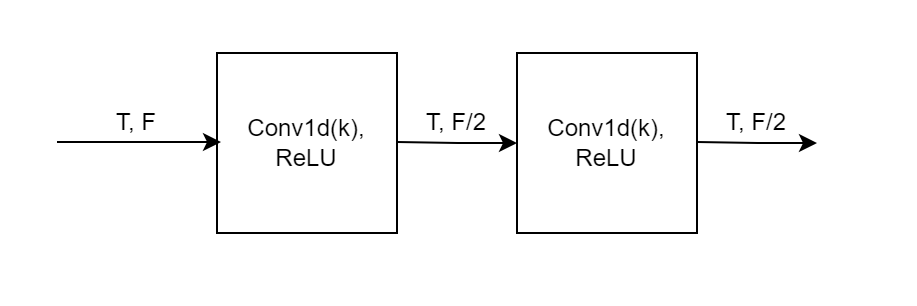}  \\
\bottomrule
\end{tabular}
\end{sc}
\end{small}
\end{center}
\vskip -0.1in
\end{table}

\begin{table}[h]
\caption{Performance of the best architectures. All models were calibrated to have the same work point where the upper limit of the $1\sigma$ confidence interval obeys $FAR_{\rm upper}\le 1\%$.}
\label{UNET-results}
\renewcommand{\arraystretch}{1.5}
\vskip 0.15in
\begin{center}
\begin{small}
\begin{tabular}{|M{2.0cm}|M{2.0cm}|M{2.0cm}|M{2.0cm}|M{2.0cm}|M{2.0cm}|}
\hline \toprule
Architecture  & $FAR \, [\%]$ & $MDR\, [\%]$ & $MBD_3$ & Number of Parameters\\ \hline
CLT  &0.77$\pm$0.30	& 12.5$\pm$1.1 & 107$\pm$5 & 0.45M \\ \hline
PLT  &0.84$\pm$0.21	& 17.4$\pm$1.2 & 157$\pm$18 & 17M \\ \hline
UNET  & 0.76$\pm$0.26
 & 18.7$\pm$1.5
 &  195$\pm$41
& 153M \\ \hline
\end{tabular}
\end{small}
\end{center}
\vskip -0.1in
\end{table}

\newpage

\section{Tokenization Specifications}
\label{appendix: tokenization}

\subsection{CLT Tokenization}
Each NVMe command, $c$, passes through the CLT tokenization scheme, as follows:
\begin{itemize}
    \item $\delta t(c) \equiv \min(round(\log((timestamp(c)-timestamp(c-1))\cdot10^5 + 1)), 15)$.
    \item $size\equiv\min(\lfloor\log(\frac{S}{512})\rfloor, 12)$. Where S is the size of the command in bytes. Since 4 bits are assigned, the remaining 3 bins (13, 14, 15) are assigned to particularly common size values: 512K, 128K, and 16K.
    \item The command's opcode ($R$/$W$) is assigned with 1 bit.
    \item The three auxiliary attributes - WAR, RAR, and RAW - are assigned 1 bit each, denoting whether their value for this command is greater than 0.
    \item The last (minor) 21 bits of the offset are dropped (this represents a 2MB resolution) and the next 2 bits are taken as $offset_{lsb}$.
    \item The first (major) 4 bits of the offset are taken as $offset_{msb}$.
\end{itemize}
In total, these attributes are spread across two tokens, where each token is composed of 9 bits: the first token contains the $\delta t(c)$, size, and opcode attributes - and the second receives the two offset attributes and the three auxiliary attributes. Then, an additional single-bit is added to each token ($0$ to the first, $1$ to the second), denoting its index in the token pair, and creating a $2^{10} = 1024$ vocabulary size.

\subsection{PLT Tokenization}
\label{PLT_tokens_details}
In this section, we detail how we normalize some of our features. 
\subsubsection{Normalization of $\delta t$, $\Delta t$, and the $offset$}
\label{dt_Dt_offset}
Starting with the $timestamp$ related quantities, for each slice $I$ and a per-command attribute $Q$ (for example $Q=\delta t$ or $\Delta t_{W, R}$) we denote by $\langle Q \rangle_{I, w}$ the average across all commands within a slice performed with the per-command weight $w_c$,
\begin{equation}
    \langle Q \rangle_{I, w}= \frac{\sum_{c\in I} Q_c w_c}{\sum_{c \in I}  w_c},
\end{equation}
and by $\langle \langle Q\rangle \rangle_{I, w}$ the exponentially back averaged of $Q_{I, w}$ across slices, defined recursively by $\langle \langle Q\rangle \rangle_{I, w} = \alpha \langle \langle Q\rangle \rangle_{I-1, w} + (1-\alpha)\langle Q\rangle_{I, w}$. In this work, we choose the diminishing factor $\alpha=0.8$, which means the effective memory of the average is $\sim 1/\log \alpha^{-1}\sim 4.48$ slices. 
For featurization, we define the following normalized quantities.
\begin{eqnarray}    \overline{offset}_I&\equiv&\left(offset - \langle offset \rangle_{I, size^2}\right)/\sqrt{\langle offset^2\rangle_{I, size^2}},\\
    \overline{\delta t}_I&\equiv&\delta t/\langle\langle \delta t\rangle\rangle_{I, size^2},\\
    \overline{\Delta t}_I&\equiv&\Delta t/\left(10\langle\langle \delta t\rangle\rangle_{I, OV^2}\right).
\end{eqnarray}
The purpose of the $\delta t$ normalization is to create an embedding that is invariant to CPU speed while normalizing the $size$ is aimed to achieve a disk size invariant embedding. 

\subsubsection{Normalization of per-token IO size and IO byte}
\label{fractions}
As mentioned in \cref{PLT_tokenization} we concatenate to the histogram features of \cref{tab: histograms} a set of 9 features associated with the number and volume of commands in a patch token. We describe their normalization in \cref{tab: UNET_embedding_simple_byVolume} and \cref{tab: UNET_embedding_simple_byCommand}. Here the normalization factors $v_0$ and $n_0$ are the designed token width in the \texttt{ByVolume} and \texttt{ByCommand} version models and were chosen as $v_0=50MB$ and $n_0=250$.

\newpage

\begin{table}[h]
\caption{The details of the last 9 features participating in the token embedding for the \texttt{ByVolume} PLT.}
\label{tab: UNET_embedding_simple_byVolume}
\renewcommand{\arraystretch}{1.5}
\vskip 0.15in
\begin{center}
\begin{small}
% \begin{sc}
\begin{tabular}{|c|c|c|}
\toprule
    Feature & Normalization factor& Number of features\\ \hline
    Patch volume & $v_0$ & 1\\
    Patch read and write volume & $v_0$ & 2\\
    Sum of $OV_{WAR}$ and $OV_{RAR}$ in a patch & $v_0$ & 2\\
    Number of read/write commands in a patch & Number of commands in a patch & 2\\     Number of $WAR$ and $RAR$ commands in a patch & Number of commands in a patch  & 2\\
\bottomrule
\end{tabular}
% \end{sc}
\end{small}
\end{center}
\vskip -0.1in
\end{table}
\begin{table}[h]
\caption{The details of the last 9 features participating in the token embedding for the \texttt{ByCommand} PLT.}
\label{tab: UNET_embedding_simple_byCommand}
\renewcommand{\arraystretch}{1.5}
\vskip 0.15in
\begin{center}
\begin{small}
\begin{tabular}{|c|c|c|}
\toprule
    Feature & Normalization factor& Number of features\\ \hline
    Patch number of commands & $n_0$ & 1\\
    Patch read and write number of commands & $n_0$ & 2\\
    Number of $WAR$ and $RAR$ commands in a patch & $n_0$ & 2\\
    The volume of read/write commands in a patch & Patch volume & 2\\     
    Sum of $OV_{WAR}$ and $OV_{RAR}$ in a patch & Patch volume & 2\\
\bottomrule
\end{tabular}
\end{small}
\end{center}
\vskip -0.1in
\end{table}

\newpage

\section{Data Specifications}
\label{appendix: data}

\subsection{Specification of the data set used}
\label{CLEAR}

Our data set was collected with a data collection, verification, and labeling system, that we developed in-house. Additional information on the data set is brought in \cref{clear_overview_table}.
In particular, we ran 49 ransomware variants list in \cref{rw_types}.
We also ran 17 types of benign software that are presented in \cref{benign_types}.

These ran on virtualization of stand-alone PCs with either a 100GB SSD whose disk usage was up to 95\%, or a 512GB SSD disk with a disk usage of 49\%-60\%. The OS was a Windows 10h22 home edition, with two CPUs: Intel(R) Core(TM) i9-10900F and i9-11900K, and with 16GB RAM. In addition, we placed different types of victim files on the virtual disk, including the Napier-Small repository (157GB) and the Napier-tiny repository (17.8GB) \cite{Napier}, some from an in-house generic user files repository, and 57GB stored in $\sim 65K$ png files from the DiffusionBM-2M data set repository. Finally, because we saw that disk indexing can be a CPU-heavy process and can significantly change the NVMe sequence, the data was generated in two configurations: where the indexing processes are either turned on or off. In total, we collected data from 6 such configurations. Each of our traces reflects up to approximately $1000$ seconds of operation time, with additional variability introduced via varying the launch time of ransomware.
The description of the images on which we ran the workloads is brought in \cref{images_info_table}.

\begin{table}[h]
\caption{Data set Overview}
\label{clear_overview_table}
\renewcommand{\arraystretch}{1.5}
\vskip 0.15in
\begin{center}
\begin{small}
\begin{tabular}{|c|c|c|c|c|c|c|M{2.0cm}|M{2.0cm}|}
\hline
Label & \multicolumn{3}{c|}{Number of Recordings} & \multicolumn{3}{c|}{Volume $[TiB]$} & \multirow{2}{2cm}{\qquad Hours} & \multirow{2}{2cm}{No. of Commands} \\ \cline{2-7}
& Train & Test & Total & Train & Test & Total & & \\ \hline
Ransomware & 971 & 493 & 1,464 & 89.6 & 45.05 & 134.65 & 402.7 & 3,896,418,082 \\ \hline
Benign & 882 & 379 & 1,261 & 29.4 & 12.96 & 42.36 & 1930 & 1,339,832,788 \\ \hline
Total & 1,853 & 872 & 2,725 & 119 & 58.01 & 177.01 & 2332.7 & 5,236,250,870 \\ \hline
\end{tabular}
\end{small}
\end{center}
\vskip -0.1in
\end{table}

\begin{table}[h]
\caption{Ransomware families in our data set}
\label{rw_types}
\renewcommand{\arraystretch}{1.5}
\vskip 0.15in
\begin{center}
\begin{small}
\begin{tabular}{|c|M{2.0cm}|} \hline
Ransomware Families & No. of Variant Streams per Family \\ \hline
Sodinokibi & 14 \\ \hline
LockBit & 9 \\ \hline
BlackMatter & 8 \\ \hline
Hive, Thanos & 7 \\ \hline
AvosLocker & 6 \\ \hline
Maze, BlackBasta, TimeTime, Babuk & 5 \\ \hline
Mespinoza, RagnarLocker, GlobeImposter & 4 \\ \hline
Play, Karma, Lorenz, Diavol & 3 \\ \hline
Sugar, WannaCry, MedusaLocker, Royal, & \multirow{3}{*}{2} \\ ViceSociety, Stop, Cuba, Rook, & \\ Conti, BianLian & \\ \hline
Neshta, CryLock, Zeppelin, Ransomware.Makop, & \multirow{6}{*}{1} \\ Alkhal, Clop, LIKEAHORSE, Teslarvng, & \\ ATOMSILO, Ransomware.Koxic, Phobos, RansomEXX, & \\ RanzyLocker, Nefilim, MRAC, Intercobros, & \\ HelloXD, DECAF, MountLocker, BlackOut, & \\ Cerber, DarkSide & \\ \hline
\end{tabular}
\end{small}
\end{center}
\vskip -0.1in
\end{table}

\begin{table}[h]
\caption{Benign SW workload types. Additional variability was introduced by using more than a single parameter for certain SWs. For example, the files read, written, deleted, archived, and encrypted were varied.}
\label{benign_types}
\renewcommand{\arraystretch}{1.5}
\vskip 0.15in
\begin{center}
\begin{small}
\begin{tabular}{|c|c|} 
\hline
Workload Type & Specific Application \\ \hline
Archiving & 7z, winRAR \\ \hline
Encryption & AESCrypt \\ \hline
Deletion & SDelete, fsutil \\ \hline
\multirow{10}{*}{Other disk accesses} & download and install apps from the internet \\
& git clone \\
& read and write files from disk \\
& conda install, pip install \\
& Windows update \\
& Compilation \\
& Document editing \\
& Web surfing \\
& Application downloading and installing \\
& Disk populating \\
\hline
\end{tabular}
\end{small}
\end{center}
\vskip -0.1in
\end{table}

\begin{table}[h]
\caption{Disk images}
\label{images_info_table}
\renewcommand{\arraystretch}{1.5}
\vskip 0.15in
\begin{center}
\begin{small}
\begin{tabular}{|c|M{9cm}|c|c|} \hline
SSD Volume $[GB]$ & Victim File Sources & Disk Occupancy & No. of Images \\ \hline
100 & \multirow{2}{9cm}{Napier, 65K png files from the
DiffusionBM-2M data set repository, a user file collection} & 10$\%$ - 95$\%$ & \multirow{2}{*}{3} \\
512 & & 50$\%$ - 60$\%$ & \\ \hline
\end{tabular}
\end{small}
\end{center}
\vskip -0.1in
\end{table}

\subsection{Details for Clustering and Robustness Analysis}
\label{clusters}
For each experiment of the leave-one-out cross-validation\cref{robustness}, the data set was partitioned into three folds.
Each fold represents the data used for an out-of-distribution test set and is presented in \cref{cluster_folds_table}. Each ransomware variant is named in the format \texttt{FAMILY}\_\texttt{ABCD} stating the ransomware family name and the first four characters of the SHA-256 hash commonly used as the ransomware signifier.

The TLSH distance chosen for the clustering process was 100 (every distance between 50 to 100 yielded the same clusters).
In rows where more than one cluster (of the same size) appears, the different clusters are distinguished and separated from one another with round brackets.

\newpage

\begin{table}[h]
\caption{Ransomware variant divided into clusters per fold.}
\label{cluster_folds_table}
\renewcommand{\arraystretch}{1.5}
\vskip 0.15in
\begin{center}
\begin{small}
\begin{tabular}{|M{10cm}|M{1.5cm}|M{1.5cm}|M{0.5cm}|} \hline
Ransomware Variant Clusters & Cluster Size & Number of Clusters & Fold\\ \hline
AvosLocker\_43b7,  Ransomare.Koxic\_7a5e,  Mespinoza\_44f1,  AvosLocker\_fb54, 
Mespinoza\_7c77,  Mespinoza\_f602,  Mespinoza\_0433,  Neshta\_9317,
Cerber\_078d,  AvosLocker\_f810,  AvosLocker\_6cc5,  AvosLocker\_c0a4, 
AvosLocker\_84d9,  Conti\_24ac,  RansomEXX\_fa28,  BlackBasta\_2558,
Lorenz\_1264,  Lorenz\_a0cc,  Lorenz\_edc2,  Teslarvng\_bb91
& 20 & 1 &  \\ \cline{1-3}
TimeTime\_5ee8, TimeTime\_972e, TimeTime\_b599, 
TimeTime\_b722, Royal\_44f5, Royal\_d9be, 
& 6 & 1 & \multirow{3}{*}{1} \\ \cline{1-3}
Babuk\_eb18, Babuk\_ca0d, Babuk\_575c, Babuk\_a522, Babuk\_77c7 & 5 & 1 &  \\ \cline{1-3}
Karma\_4dec, Karma\_3462, Karma\_84d2 & 3 & 1 &  \\ \cline{1-3}
Cuba\_21ac, BlackBasta\_723d & 2 & 1 & \\ \cline{1-3}
ViceSociety\_HelloKitty\_fa72, Alkhal\_7a31, ATOMSILO\_d9f7, Stop\_0d50, Sugar\_09ad,
Maze\_4263, MRAC\_768c, Phobos\_265d, WannaCry\_be22, Play\_952f
& 1 & 10 &  \\ \toprule\bottomrule

Sodinokibi\_fd16, Sodinokibi\_b992, Sodinokibi\_0441, Sodinokibi\_6834, Sodinokibi\_20d4,
Sodinokibi\_9df3, Sodinokibi\_de20, Sodinokibi\_9b11, Sodinokibi\_cb4a, Sodinokibi\_2f00,
Sodinokibi\_9437, Sodinokibi\_7c8c, Sodinokibi\_db59, Sodinokibi\_3b0c 
& 14 & 1 & \\ \cline{1-3}
BlackOut\_ee13, Diavol\_b3da, Diavol\_7945, Diavol\_2723, 
Play\_006a, ViceSociety\_HelloKitty\_c249, RanzyLocker\_0db6, Play\_dd10 
& 8 & 1 & \multirow{3}{*}{2}  \\ \cline{1-3}
Thanos\_caf8, Thanos\_8141, Thanos\_66ed, Thanos\_4852, 
Thanos\_6e6b, Thanos\_cbdb, Thanos\_d29a
& 7 & 1 & \\ \cline{1-3}
MountLocker\_00ed, MedusaLocker\_f5fb, MedusaLocker\_c2a0 & 3 & 1 &  \\ \cline{1-3}
$\left(\right.$BlackBasta\_df5b, GlobeImposter\_185f$\left)\right.$, $\left(\right.$Rook\_c2d4, LockBit\_f2da$\left)\right.$  & 2 & 2 &  \\ \cline{1-3}
WannaCry\_d103, Cuba\_521c, Zeppelin\_824a, Sugar\_1d4f, DECAF\_a471, 
Ransomare.Makop\_a617, Stop\_59d0, Conti\_53b1, Intercobros\_ade5, HelloXD\_903c
& 1 & 10 &  \\ 

\toprule\bottomrule

Hive\_9e9f, Hive\_0320, Hive\_a45c, Nefilim\_fb3f, Hive\_460b, 
Hive\_0302, Hive\_dfa5, BianLian\_eaf5, Hive\_45fe, BianLian\_46d3
& 10 & 1 &  \\ \cline{1-3}
$\left(\right.$ BlackMatter\_8ead, BlackMatter\_22d7, BlackMatter\_e4fd, BlackMatter\_c6e2, 
BlackMatter\_730f, BlackMatter\_b824, BlackMatter\_2aad, BlackMatter\_5da8$\left. \right)$,
$\left(\right.$LockBit\_bdc2, LockBit\_e216, LockBit\_4bb1, LockBit\_0545, 
LockBit\_acad, LockBit\_7340, LockBit\_dd8f, LockBit\_786a$\left.\right)$
& 8 & 2 &  \\ \cline{1-3}
$\left(\right.$Maze\_e8a0, Maze\_3885, Maze\_6a22$\left.\right)$, 
$\left(\right.$GlobeImposter\_e6fa, GlobeImposter\_39f5, GlobeImposter\_70fa $\left.\right)$
& 3 & 2 & 3 \\ \cline{1-3}
$\left(\right.$RagnarLocker\_10f9, RagnarLocker\_5469$\left)\right.$, $\left(\right.$BlackBasta\_e281, BlackBasta\_c4c8$\left)\right.$ & 2 & 2 &  \\ \cline{1-3}
RagnarLocker\_afab, CryLock\_4a47, RagnarLocker\_041f, LIKEAHORSE\_6d2e, 
DarkSide\_f3f2, Rook\_f87b, TimeTime\_c535, Clop\_bc1f, Maze\_4e25
& 1 & 9 &  \\ \hline

\end{tabular}
\end{small}
\end{center}
\end{table}

\clearpage

\section{Hardware Specifications}
\label{appendix: HW}
Here we provide detailed calculations of the hardware requirements we presented in the main paper. 
First we calculate the number of multiplication required by each model for its forward pass. Other operations such as additions, exponentiation, etc. that are part of the forward pass as well, are much fewer and cheaper to implement in hardware and are therefore omitted here. We perform estimates for both the CLT and the PLT in the following section.
\subsection{Calculations for the CLT}
\label{CLT_HW_calcs}
Following the model details found in \cref{clt-hyperparams}, the CLT model architecture includes
\begin{itemize}
    \item Embedding table of size $1024\times128=131072$
    \item Positional encoding of size $500\times128=64000$
    \item 3 self-attention encoder layers, each containing
    \begin{itemize}
        \item $W_Q,\,W_K,\,W_V$ matrices, each of size $128\times128=16384$ and total size of $3\cdot16384=49152$
        \item Two fully connected layers, each of size $128\times128=16384$ and total size of $2\cdot16384=32768$
        \item bias vectors, $\beta,\,\gamma$ parameters for the layer norm, each of size $128$ and total size of $4\cdot128=512$
    \end{itemize}
    \item linear projection layer of size $128+1=129$
    \item $2\times1$ convolution kernel of size $2$
\end{itemize}
The total number of parameters in the model is thus
$$131072+64000+3(49152+32768+512)+129+2\simeq442K$$

The number of multiplications can similarly be derived for N - input vector of length 500
\begin{itemize}
    \item For the self-attention we get
    \begin{itemize}
        \item $NW_Q=Q,\,NW_K=K,\,NW_V=V$ is $500\times128^2\simeq8M$ each and $8M\cdot3\cdot3=72M$ in total
        \item $QK^T=X,\,XV$ is $500^2\times128=32M$ each and $32M\cdot2\cdot3=192M$ in total
    \end{itemize}
    \item For the fully connected layers we get
    \begin{itemize}
        \item $500\times128^2=8M$ each and $8M\cdot2\cdot3=48M$ in total
    \end{itemize}
    \item For the final linear projection we get $500\times128=64K$
\end{itemize}
The total number of multiplications in a forward pass is thus
$$72M+192M+48M+64K\simeq312M$$

Finally, we note that since the CLT model works on by-command slices of 250 commands each, we may want to make sure it achieves the SSD throughput in random IOPS as well. The calculation is quite similar. For 312M multiplications per forward pass, a 300MHz hardware clock, 256 multipliers, and 250 commands per slice we get
$$\frac{250\cdot300\text{MHz}\cdot256}{312\text{M}}\simeq60\text{K}$$ IOPS
Well above the spec of even the fastest SSD devices currently on the market.

subsection{Calculations for the PLT}
\label{PLT_HW_calcs}
Following the model details found in \cref{plt-hyperparams}, the PLT model architecture includes
\begin{itemize}
    \item Embedding matrix of size $181\times512=92672$
    \item Positional encoding of size $100\times512=51200$
    \item 3 self-attention encoder layers, each containing
    \begin{itemize}
        \item $W_Q,\,W_K,\,W_V$ matrices, each of size $512\times512=262144$ and total size of $3\cdot262144=786432$.
        \item Two fully connected layers, each of size $512\times2048=1048576$ and total size of $2\cdot1048576=2097152$.
        \item bias vectors, $\beta,\,\gamma$ parameters for the layer norm, each of size $512$ and total size of $4\cdot512=2048$.
    \end{itemize}
    \item linear projection layer of size $2\times512+2=1026$
\end{itemize}
The total number of parameters in the model is thus
$$92672+51200+6(786432+2097152+2048)+1026\simeq17.5M$$

The number of multiplications can similarly be derived for N - input vector of length 100
\begin{itemize}
    \item For the first embedding we get $181\times512\times100=9M$
    \item For the self-attention we get
    \begin{itemize}
        \item $NW_Q=Q,\,NW_K=K,\,NW_V=V$ is $100\times512^2\simeq26.2M$ each and $26.2M\cdot6\cdot3=472M$ in total
        \item $QK^T=X,\,XV$ is $100^2\times512=5M$ each and $5M\cdot2\cdot6=60M$ in total
    \end{itemize}
    \item For the fully connected layers we get
    \begin{itemize}
        \item $100\times512\times2048=105M$ each and $105M\cdot2\cdot6=1258M$ in total
    \end{itemize}
    \item For the final linear projection we get $100\times512\times2=102K$
\end{itemize}
The total number of multiplications in a forward pass is thus
$$9M+472M+60M+1258M+102K\simeq1.8G$$

\end{document}